\documentclass[letterpaper, 10 pt, conference]{ieeeconf}  

\IEEEoverridecommandlockouts                              

\usepackage{graphicx} 
\usepackage{amsmath} 
\usepackage{amssymb}  
\usepackage{bm}
\usepackage{subfigure}
\usepackage{color}
\usepackage{cite}
\usepackage[linesnumbered,ruled]{algorithm2e}
\usepackage{hyperref}
\usepackage{float}
\usepackage{booktabs}
\usepackage{multirow}
\usepackage[utf8]{inputenc}
\hypersetup{
	colorlinks=true,
	linkcolor=blue,
	filecolor=magenta,      
	urlcolor=cyan,
}
\usepackage{pifont} 
\newcommand{\xmark}{\text{\ding{55}}}
\newcommand{\etal}{\textit{et al.}}
\newcommand{\ie}{\textit{i.e.,} }
\newcommand{\eg}{\textit{e.g.,} }

\def\figureautorefname~#1\null{Fig.~#1\null}
\def\equationautorefname~#1\null{Eq.~(#1)\null}
\newcommand{\Autoref}[1]{%
	\begingroup%
	\def\algorithmautorefname~##1\null{Algorithm~##1\null}%
	\autoref{#1}%
	\endgroup%
}

\title{\LARGE \bf
	Towards Robust Visual Tracking for Unmanned Aerial Vehicle with Tri-Attentional Correlation Filters
}

\author{Yujie He$^{1}$, Changhong Fu$^{1,*}$, Fuling Lin$^{1}$, Yiming Li$^{1}$, and Peng Lu$^{2}$
	\thanks{*Corresponding author}
	\thanks{$^{1}$Yujie He,
		Changhong Fu, Fuling Lin, and Yiming Li are with the School of Mechanical Engineering, Tongji University, 201804 Shanghai, China
		{\tt\small changhongfu@tongji.edu.cn}}%
	\thanks{$^{2}$Peng Lu is with the Adaptive Robotic Controls Lab (ArcLab), Hong Kong Polytechnic University (PolyU), Hong Kong, China
		{\tt\small peng.lu@polyu.edu.hk}}%
}

\begin{document}

	\maketitle
	\thispagestyle{empty}
	\pagestyle{empty}

	\begin{abstract}
		
		Object tracking has been broadly applied in unmanned aerial vehicle (UAV) tasks in recent years. However, existing algorithms still face difficulties such as partial occlusion, clutter background, and other challenging visual factors. 
		Inspired by the cutting-edge attention mechanisms, a novel object tracking framework is proposed to leverage multi-level visual attention. 
		Three primary attention, \ie contextual attention, dimensional attention, and spatiotemporal attention, are integrated into the training and detection stages of correlation filter-based tracking pipeline. Therefore, the proposed tracker is equipped with robust discriminative power against challenging factors while maintaining high operational efficiency in UAV scenarios.
		Quantitative and qualitative experiments on two well-known benchmarks with 173 challenging UAV video sequences demonstrate the effectiveness of the proposed framework. The proposed tracking algorithm favorably outperforms 12 state-of-the-art methods, yielding 4.8\% relative gain in UAVDT and 8.2\% relative gain in UAV123@10fps against the baseline tracker while operating at the speed of $\sim$28 frames per second.
	\end{abstract}
	
	\section{Introduction}\label{sec:intro}
	
	Visual object tracking plays an essential role in unmanned aerial vehicle (UAV) tasks, including target following~\cite{Cheng2017IROS}, flying vehicle tracking~\cite{Fu2014ICRA}, and autonomous landing~\cite{Lin2017AutonomousRobots}. However, most existing trackers remain vulnerable under challenging environmental conditions. In UAV tracking, the complex working conditions continuously lead to different types of object appearance variations, such as partial occlusion and background clutter, which severely degrade the overall tracking accuracy and robustness.
	
	Under the hard constraints of a real-time vision-based UAV system, an ideal tracker should be efficient to gain more computing capability for sensor fusion, high-level control, \textit{etc}.
	Compared with other discriminative tracking methods, correlation filter (CF)-based trackers assume circulant shifts of the object sample for filter training. 
	Efficient element-wise operations in the Fourier domain can guarantee real-time performance, which is favorably suitable for UAV tracking applications.
	Thus, CF-based tracking methods have been extensively studied in recent years~\cite{Wang2018AAAI, Li2018CVPR, Dai2019CVPR, Huang2019ICCV, Lin2020ICRA, Fu2020TGRS}. 
	In this process, however, the circulant shifting operation and extended search region introduce synthetic samples and unwarranted noise separately~\cite{Danelljan2015ICCV}. 
	Thus, the learned filters are easily corrupted and can lead to unrecoverable drift when background clutter, similar objects, or other challenging factors occurred.
	
	\begin{figure}[!t]
		\centering
		\includegraphics[width=0.48\textwidth]{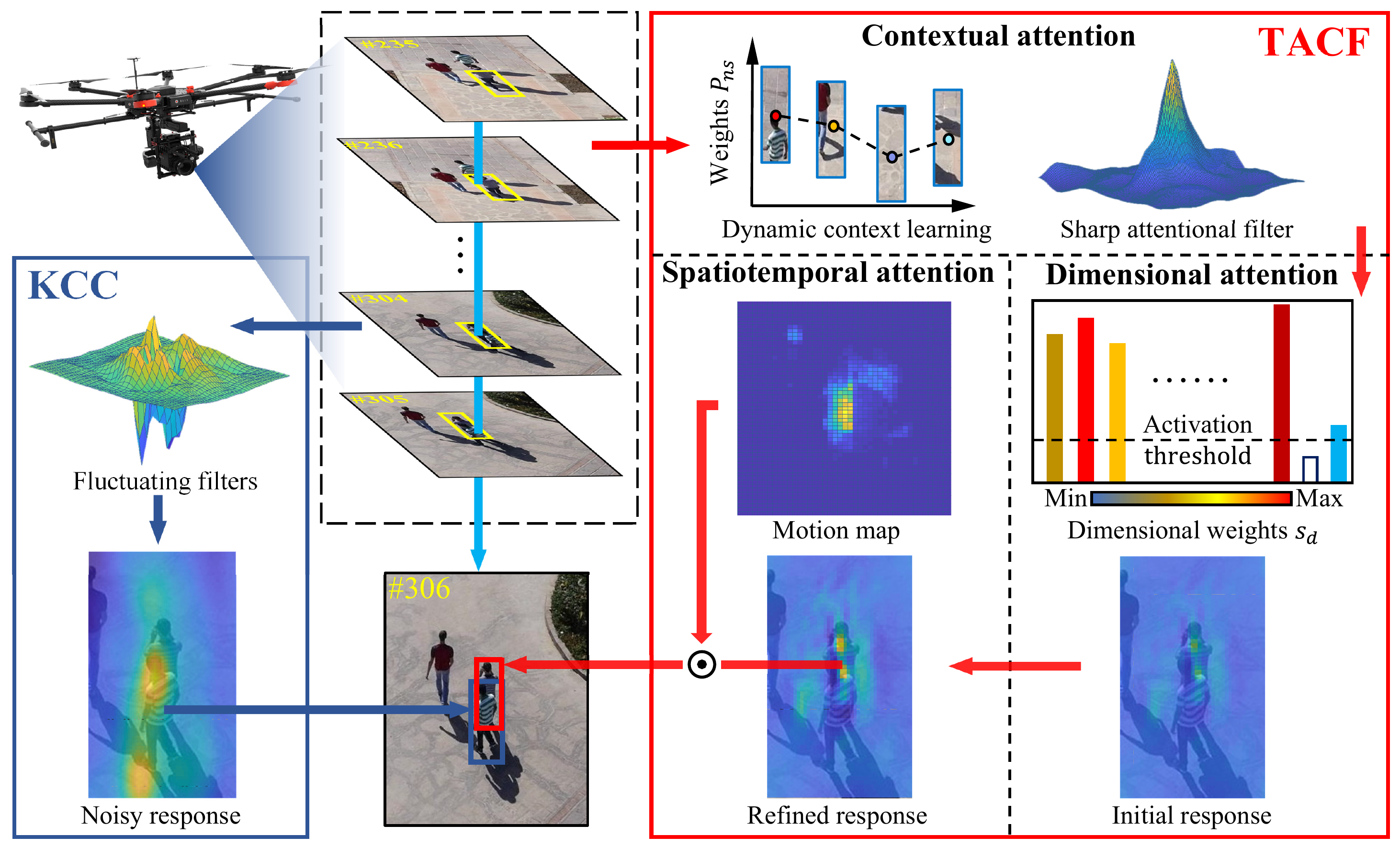}
		\caption{Comparison between baseline KCC tracker~\cite{Wang2018AAAI} and the proposed TACF tracker. In UAV tracking, TACF incorporates the contextual information into the filter training stage, so that the attentional filters can sense the varying appearance of the object and environment simultaneously. After that, both the dimensional and spatiotemporal attention of the response maps are utilized to enhance the overall map, aiming to repress the noise and improve the tracking performance better. In contrast, the KCC tracker only uses the sample information of the current frame, which makes the filter susceptible to the appearance changes.}
		\label{fig:innovation}
	\end{figure}
	
	Some works have attempted to tackle background clutter or similar object by applying context learning. 
	Background-aware CF (BACF)~\cite{Galoogahi2017CVPR} is proposed to extract more negative samples from the larger background by constructing a cropping matrix. 
	M. Mueller~\etal~\cite{Mueller2017CVPR} managed to expand the receptive fields of filters by utilizing context information. 
	Although these methods are capable of sensing the surrounding environment, their excessive samples lack a reasonable attention mechanism. 
	Therefore, they are not able to employ the limited computational resources to the parts that may interfere with the tracked object.
	
	Besides, the CF-based tracking method mostly ignores the potential information about the multi-dimension response maps itself or from consecutive frames.
	The responses generated by correlation filters indicate the importance of some particular pixels in the different feature dimensions and locations. 
	What is more, the temporal cues from video data can provide consistency constraints to improve tracking accuracy further. 
	Some algorithms~\cite{Lukezic2017CVPR, Dai2019CVPR} applied the color-based or iterated spatial masks to capture useful information related to tracked objects.
	However, these methods can hardly cope with the occurrence of cluttered backgrounds and partial occlusion because they merely exploited few levels of information collaboratively and almost consider only a single frame.
	
	Inspired by the human perception system, the effectiveness of attention mechanisms become increasingly affirmed by many fields, especially in computer vision for robotics~\cite{Hu2018CVPR, Woo2018ECCV, Chen2018RAL, Fu2019CVPR}. 
	By intuitively introducing the multi-level visual attention into the training and detection stages of the CF-based framework, the tracker can be enhanced with a comprehensive attention framework. The object-oriented capability can increase the discriminative power against partial occlusion, background clutter, and other challenging factors, which often appear in UAV tracking scenes.
	
	In this work, a novel tracking framework with tri-attention correlation filters is proposed to utilize the importance of different parts more comprehensively for more adaptive filter training and response generation. 
	With contextual attention strategy, this work manages to exploit the dynamic responses by assigning different penalty factors to context patches, so that the filters can increase perception to the environment and discriminative capability to the object simultaneously. 
	When generating response maps, both feature dimension-based attention and spatiotemporal attention are proposed to enhance the quality by relocating the focus on the tracked object-oriented information.
	As a result, a novel tri-attention correlation filter, \ie TACF tracker, is proposed to enhance the robustness against partial occlusion, clutter background, and other challenging factors in UAV tracking.
	The comparison between the proposed tracking method with the baseline tracker is depicted in \autoref{fig:innovation}. TACF shows better performance than the baseline, especially when the target is partially occluded or shares similar visual cues with surrounding objects.
	
	The main contributions of this work are listed as follows:
	\begin{itemize}
		\item A unified tri-attention framework to leverage multi-level visual information, including contextual, spatiotemporal, and dimensional attention, is proposed to improve tracking performance in both robustness and efficiency.
		
		\item By introducing a novel criterion to assess the quality of the contextual response map, the contextual attention strategy integrates dynamically varying environmental information into the filter training stage.
		
		\item The spatiotemporal and dimensional attention modules are applied to exploit both the internal and external associations of consecutive frames. Therefore, the methods can rule out unwarranted noise along the tracking process by allocating different levels of attention to response maps dynamically.
		
		\item Extensive evaluations have conducted on two well-known UAV tracking benchmarks with 173 challenging image sequences. The results have demonstrated that the presented TACF tracker outperforms the other 12 state-of-the-art trackers while operating at the speed of $\sim$28 frames per second (FPS).
	\end{itemize}
	
	\section{Related works}
	\label{sec:related}
	
	\subsection{Tracking with correlation filters}
	Exploiting CF for object tracking started with the method called the minimum output sum of squared error, \ie MOSSE tracker~\cite{Bolme2010CVPR}. The tracker is constructed and trained using gray-scale samples in the frequency domain for efficiency. Afterward, a variety of works built upon the framework improve the performance by combining multiple features~\cite{Danelljan2014CVPR}, scale estimation~\cite{Danelljan2014BMVC}, kernel trick~\cite{Henriques2015TPAMI}, context learning~\cite{Mueller2017CVPR, Galoogahi2017CVPR}, filter weighing~\cite{Danelljan2015ICCV, Dai2019CVPR, Huang2019ICCV, Fu2020NCAA},  or end-to-end neural network architecture~\cite{Valmadre2017CVPR, Zhu2018CVPR, Choi2017CVPR}. Kernel cross-correlator (KCC)~\cite{Wang2018AAAI} provides a novel solution for the CF-based framework with high expandability and brief formulation. However, the lack of context information and further refinement of generated responses limit the tracker's ability to discriminate the object in diversified environments.
	
	Other CF-based methods focus on utilizing features extracted from the convolutional neural network (CNN) to obtain a more comprehensive object representation. Some trackers hierarchically utilize convolutional features or propose an adaptive fusion approach to improve the encoding ability of the model~\cite{Ma2015ICCV, Danelljan2016ECCV, Danelljan2017CVPR, Danelljan2019PRL}. 
	Although CF-based trackers have made sound progress, it is still difficult for them to achieve object tracking for UAV with high performance and efficient operation at the same time.
	
	\subsection{Tracking with attention mechanism}
	Attention, as a critical ability of the human system, enables people to focus on more useful information when processing multi-modal information.  Originated from the area of neural machine translation for modeling the contextual information, attention mechanism in machine vision has improved the success of various applications in recent years. It continues to be an omnipresent component in state-of-the-art models, such as image classification~\cite{Hu2018CVPR}, image captioning~\cite{Chen2016CVPR}, and scene segmentation~\cite{Fu2019CVPR}. In visual tracking, CNN-based methods with attention mechanisms can integrate different visual information to improve tracking accuracy. J. Choi~\etal~\cite{Choi2017CVPR} proposed an attention network to switch among different features to select the suitable tracking mode. Z. Zhu~\etal~\cite{Zhu2018CVPR} incorporated optical flow based on deep learning into the tracking pipeline. Other tracking methods~\cite{Wang2018CVPR, Pu2018NIPS, Chen2019PR, Zhu2018ECCV} used feature maps extracted from deep neural networks to select the appropriate tracking mode for better performance. However, these methods are not suitable for the complex UAV tracking scenarios, \eg partial occlusion, clutter background, and viewpoint changes. 
	
	In this work, contextual attention is first applied in the filter training stage to improve the discrimination of cluttered backgrounds and similar objects in UAV tracking. Subsequent dimensional and spatiotemporal attention in the response generation stage can refine the final tracking result, leading to higher robustness and accuracy of tracker while preserving sufficient tracking speed. 
	
	\begin{figure*}[!ht]
		\centering
		\includegraphics[width=0.96\textwidth]{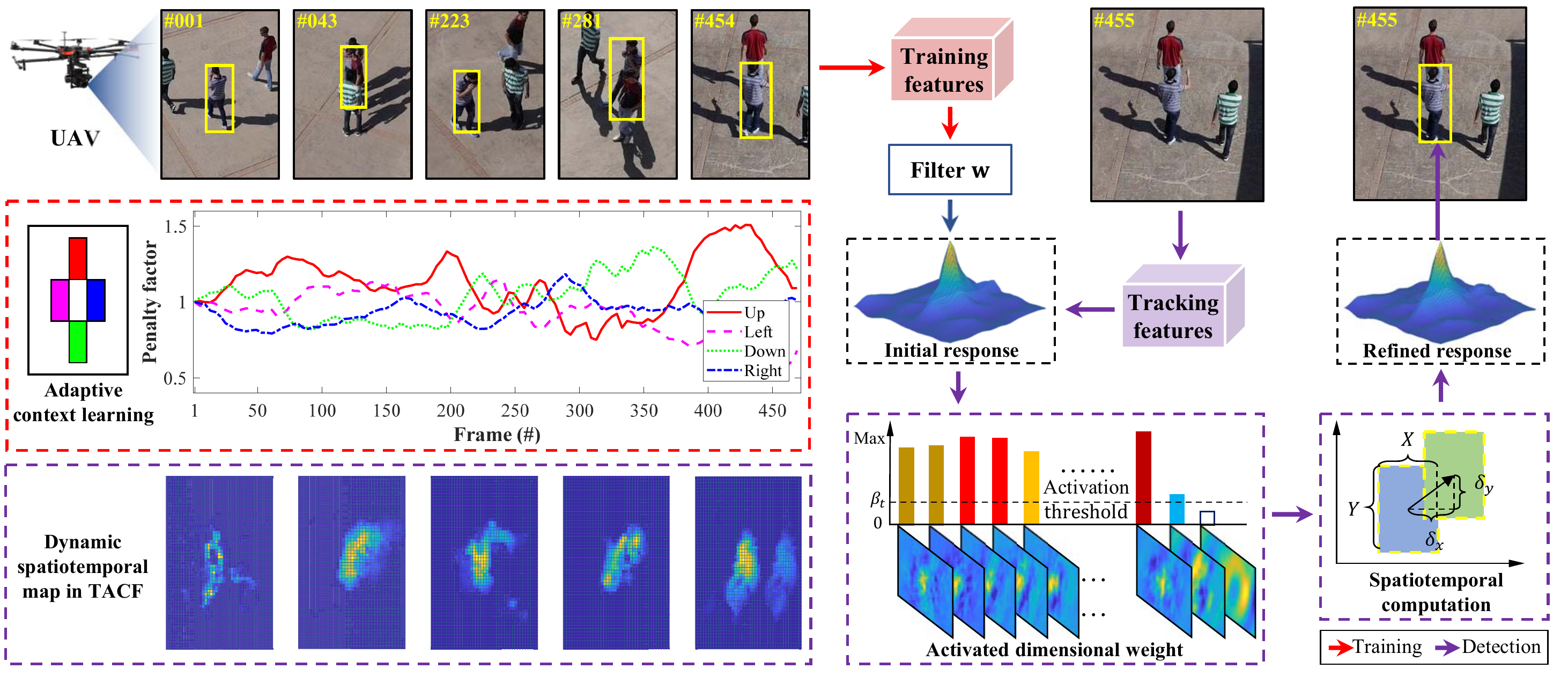}
		\caption{The main workflow of the TACF tracker. 
			The object and context patches near the object are first extracted and represented as training samples. Then, the contextual attention strategy is implemented to realize adaptive learning of the surrounding environment throughout the online tracking process, thereby notably improving the discriminative power of the tracker. After generating the response, dimensional attention and spatiotemporal attention are applied to refine the response maps and suppress noise. As a result, the proposed method can locate the tracked object efficiently and robustly.
		}
		\label{fig:main}
	\end{figure*}
	
	\section{Proposed method}
	\label{sec:proposed}
	
	This section first reviews the baseline tracker, \ie KCC~\cite{Wang2018AAAI}. 
	Then, the proposed TACF tracker is presented in a top-down way: the establishment and solution of the objective function are followed by the introduction of the tri-attention strategy. 
	The main workflow of TACF can be illustrated in \autoref{fig:main}.
	
	\subsection{Revisiting KCC}
	\label{sec:reviewKCC}
	
	Given the training and testing samples, they are denoted as column vectors $\mathbf{x}$, $\mathbf{z} \in \mathbb{R}^M$ in the subsequent derivation for clarity, which can be extended to the two-dimensional image.
	With the function $ \varphi(\cdot): \mathbb{R}^M \rightarrow \mathbb{R}^H, H \gg M $, the inner product between $\mathbf{x}$ and $\mathbf{z}_i$ can be mapped into high-dimensional space. Accordingly, the kernelized correlator between them is defined as $
	\kappa(\mathbf{x},\mathbf{z}_i) = \varphi(\mathbf{x})^\mathsf{T} \varphi (\mathbf{z}_i) \in \mathbb{R}
	$, 
	\noindent where the superscript $ ({\cdot})^{\mathsf{T}} $ denotes the transpose operation.
	Then, the sample-based vector $\mathbf{z}_i\in \mathbb{R}^M$ is computed from the test sample $\mathbf{z}$ with the transform function $ \mathcal{T}(\cdot)$ as $ \mathbf{z}_i \in \mathcal{T} (\mathbf{z})$.
	As a result, the sample-based vector set can construct the kernel vector $ \mathbf{k}^{\mathbf{xz}} = [k_{1}^{\mathbf{xz}}, \cdots, k_{n}^{\mathbf{xz}}]^{\mathsf{T}}$, 
	where $\kappa(\mathbf{x},\mathbf{z}_i)$ is denoted as $k_{i}^{\mathbf{xz}}$ for simplicity.
	Finally, the kernelized cross-correlation can encode the pattern of training
	samples into filters $ \hat{\mathbf{w}}^{*} $, and the output $ \hat{C}(\mathbf{x},\mathbf{z}) $ in frequency domain can be computed as:
	\begin{equation}
	\label{equ:kernel_correlation_output}
	\hat{C}(\mathbf{x},\mathbf{z})
	=
	\hat{\mathbf{k}}^{\mathbf{xz}} \odot \hat{\mathbf{w}}^{*}
	=
	\mathcal{F}^{-1} (\mathbf{k}^{\mathbf{xz}} \star \mathbf{w})
	\ ,
	\end{equation}
	\noindent where $\odot$ denotes element-wise product, and $ \star $ stands for cross correlation.
	The superscript 
	$(\cdot)^{*}$ 
	and 
	$\hat{\cdot}$ 
	represent complex conjugate operation and discrete Fourier transformation, \ie
	$
	\hat{\mathbf{x}} = \mathcal{F} (\mathbf{x})
	$.
	With the ideal response $ \mathbf{y} $ and learning samples $ \mathbf{x}_i $, the  objective function is formulated by minimizing  the squared error using ridge regression:
	\begin{equation}
	\label{equ:origin_func}
	\hat{\mathcal{E}}
	=
	\sum_{n=1}^{N}
	\left\Vert
	\hat{\mathbf{k}}^{\mathbf{xz}}_{n} \odot \hat{\mathbf{w}}^{*}_{n}
	-
	\hat{\mathbf{y}}
	\right\Vert_{2}^{2}
	+ 
	\lambda
	\left\Vert
	\hat{\mathbf{w}}^{*}
	\right\Vert_{2}^{2}
	\ ,
	\end{equation}
	\noindent where $ \hat{\mathbf{w}}^{*}_{n} $ is the $n$-th channel of the learned filter.
	
	Because the operations in \autoref{equ:origin_func} can be performed in element-wise,
	the corresponding $ \mathbf{w}^{*} $ can be solved independently to obtain a closed-form solution.
	However, KCC utilizes a relatively larger search area, which will bring cluttered information into filters training so that the KCC framework cannot achieve better performance by increasing distinguishing ability against real background information.
	
	\subsection{Tri-attention correlation filters framework}
	
	As reviewed in Section~\ref{sec:reviewKCC}, KCC tracker cannot fully exploit the information of surrounding contexts or enhance the critical parts what filters need to pay attention.
	Thus, attentional contextual information is introduced as negative samples to enhance the training of correlation filters, which is defined as:
	\begin{equation}
	\label{equ:context}
	\sum_{s =1}^{S} 
	\Big\Vert p_{ns} \hat{C}_n (\textbf{x}_{ns}, \textbf{x}_{ns})
	\Big\Vert_2^2
	\ ,
	\end{equation}
	\noindent where $ S $ is the number of patches extracted from the up, down, left, and right to the object.
	They are considered as hard negative samples, so the desired output is zero.
	Besides, an adaptive penalty factor $p_{ns}$ is proposed to evaluate the importance of the context patches and varies along the tracking process (a detailed explanation is in Section~\ref{subsec:ca}).
	
	Motivated by attention mechanism, spatiotemporal and dimensional attention are integrated into final response generation stage.
	Accordingly, the tri-attention correlation filters with $ N $ features taken into account can be formed by minimizing the regression target:
	\begin{equation}\label{equ:object_function}
	\begin{split}
	\hat{\mathcal{E}}(\hat{\textbf{w}}^{*})
	=
	& \sum_{n=1}^N
	\Big(
	\big\Vert \hat{C}_{n}( \textbf{x}_{n0}, \textbf{x}_{n0}) - \hat{\textbf{y}}_{n} \big\Vert_2^2 \\
	+ &
	\lambda_{1} \big\Vert \hat{\textbf{w}}_n^{*} \big\Vert_2^2
	+
	\lambda_{2} \sum_{s =1}^{S} 
	\Big\Vert p_{ns} \hat{C}_n (\textbf{x}_{ns}, \textbf{x}_{ns})
	\Big\Vert_2^2 
	\Big)
	\end{split}
	\ ,
	\end{equation}
	\noindent where $\mathbf{x}_{n0}$ and $\mathbf{x}_{ns}$ are the representation of image corresponding to the object and context by the $n$-th feature. Then, $\hat{\mathcal{E}}$ is an error measured by the correlation output of $\mathbf{x}_{n0} \in \mathbb{R}^M$ and desired output $\mathbf{y}_{n} \in \mathbb{R}^M$. 
	$ \hat{\mathbf{w}}_{n}^{*} \in \mathbb{C}^M $ denotes the correlation filter for $ n $-th feature in the Fourier domain. 
	$\lambda_{1}$ and $\lambda_{2}$ are regularization factors for the filters and context information learning.
	
	Due to the mutual independence of different features and dimensions, the objective function  $ \hat{\mathcal{E}}(\hat{\mathbf{w}}^{*}) $ in \autoref{equ:object_function} can be reformulated sub-problems $\hat{\mathcal{E}}_n$ that can be obtained as:
	\begin{equation}\label{equ:E(wn)}
	\begin{split}
	\hat{\mathcal{E}}_n
	= &
	\big\Vert \hat{C}_{n}( \textbf{x}_{n0}, \textbf{x}_{n0}) - \hat{\textbf{y}}_{n} \big\Vert_2^2
	+
	\lambda_{1} \big\Vert
	\hat{\textbf{w}}_{n}^{*} \big\Vert_2^2 \\
	&+
	\sum_{s =1}^{S} \Big\Vert P_{ns} \hat{C}_n (\textbf{x}_{ns}, \textbf{x}_{ns})\Big\Vert_2^2
	\end{split}
	\ ,
	\end{equation}
	\noindent where the regularized factor for each context patch $ P_{ns} $ can be computed as 
	$ P_{ns} = \sqrt{\lambda_2} p_{ns}, \ s = 1, \cdots, S$.
	
	By setting the first derivative of $ \hat{\mathbf{w}}_{n}^{*} $ to zero, the solution to the optimization problem \autoref{equ:E(wn)} can be calculated with element-wise operations as follows:	
	\begin{equation}
	\label{equ:model_result}
	\small
	\hat{\bf{w}}_{n}^{*} 
	= 
	\frac
	{    
		\hat{\textbf{K}}^{n0}
		\odot
		\hat{\textbf{y}}_{n}
	}
	{
		\hat{\textbf{K}}^{n0}
		\odot
		\hat{\textbf{K}}^{n0*}
		+
		\lambda_{1}
		+
		\sum\limits_{s = 1}^{S} 
		\Big(
		{P_{ns}}^2 
		\hat{\textbf{K}}^{ns}
		\odot
		\hat{\textbf{K}}^{ns *}
		\Big)
	}
	\ ,
	\end{equation}
	\noindent where the fraction operator denotes element-wise division, and $\hat{\textbf{k}}^{\textbf{x}_{n0} \textbf{x}_{n0}}$ and $\hat{\textbf{k}}^{\textbf{x}_{ns} \textbf{x}_{ns}}$ are replaced by $\hat{\textbf{K}}^{n0}$ and $\hat{\textbf{K}}^{ns}$ for clarity, respectively.
	
	For contextual patches, learning at each frame can lead to overfitting.
	Therefore, the context attention is set to update at a frequency of $ f_c $, which can further improve the overall tracking efficiency.
	When the context patches are not taken into account, \autoref{equ:model_result} can be reformulated as follows:
	\begin{equation}
	\small
	\label{equ:model_result_no_context}
	\hat{\bf{w}}_{n}^{*} 
	= 
	\frac
	{    
		\hat{\textbf{K}}^{n0}
		\odot
		\hat{\textbf{y}}_{n}
	}
	{
		\hat{\textbf{K}}^{n0}
		\odot
		\hat{\textbf{K}}^{n0*}
		+
		\lambda_{1}
	}
	\ .
	\end{equation}
	
	\subsection{Contextual attention strategy}
	\label{subsec:ca}
	
	When encountering dramatic changes in object appearance, such as occlusion or sudden illumination changes, the constant updating of the model may introduce some noisy negative samples for training, thereby reducing the quality of the filters. 
	Therefore, a novel response quality index, the peak median energy ratio (PME), is proposed for efficient response map evaluation, and give guidance for subsequent filters training, which can be calculated as follows:
	\begin{equation}
	\small
	\label{equ:PME}
	PME
	=
	\frac{\left|R_{\max} - R_{\mathrm{med} }\right|^{2}}
	{\mathrm{mean}\left[
		\sum\limits_{x = 1}^{W}
		\sum\limits_{y = 1}^{H}
		\left(R_{(x, y)}-R_{\mathrm{med}} \right)^{2}
		\right]}
	\ ,
	\end{equation}
	\noindent where $ R_{\max} $ and $ R_{\mathrm{med} } $ denote the maximum and median score separately,
	and $ R_{(x, y)} $ is pixel-wise value in the map. 
	Accordingly, the difference between $R_{\max}$ and $R_{\mathrm{med}} $ can respond to the sharpness of the highest peak. 
	The denominator $ R_{(x,y)} - R_{\mathrm{med}} $ can be applied to measure the overall smoothness.
	Therefore, the sharper peaks and smoother fluctuations in response maps can lead to a higher $R_{\max}$ and smaller $R_{\mathrm{med}} $, so that a higher PME value is obtained to indicate high quality in the response.
	Accordingly, the challenging factor of surrounding patches against the object patch is defined as follows:
	\begin{equation}
	\label{equ:challenging}
	c_s
	=
	\frac{ PME(R_s) }
	{ PME(R_0) }
	\ ,
	\end{equation}
	\noindent where $ R_0 $ and $ R_s $ denote as the response generated from the object and context patch.
	As a result, the penalty factor to each context patch $s$ can be defined as follows:
	\begin{equation}
	\small
	\label{equ:penalty_factor}
	p_s
	=
	\frac{c_s^2}{\sum\limits_{s=1}^{S} c_s^2}
	\ .
	\end{equation}
	
	As illustrated in \autoref{fig:main}, the PME index exhibits a high sensitivity to the dramatic appearance change.
	When challenging issues addressed, the quality of the response map is restored, and the index will recover to a reasonable level.
	
	
	\subsection{Dimensional attention strategy}
	\label{subsec:da}
	
	The various dimensions of the response generated by the different features can be considered as the capture of distinct sub-characteristic of the tracked object.
	By exploring the interdependence between different dimensions, the semantics of feature expression can be enhanced, and the activation strategy can be used to improve the tracking accuracy further.
	Given the multi-dimension response $ R \in \mathbb{R}^{H \times W \times D} $, the weight of different dimensions is computed as follows:
	\begin{equation}
	z_{d}
	=
	F_a ( R_d )
	=
	\frac{1}{H W} \sum_{i=1,j=1}^{H,W} R_d(i, j)
	+
	\max ( R_d )
	\ ,
	\end{equation}
	\noindent where the output of function $F_a(\cdot)$ can be interpreted as an abstract representation of a particular dimension.
	It is common to use this information in prior feature engineering work.
	In this paper, global average and maximum operations are selected for fast calculation and accurate evaluation of the response for each dimension.
	
	To fully capture inter-dimension dependencies, a consequent activation is utilized to employ a simple attention mechanism. 
	The operation enhances the ability to learn the non-mutual-exclusive associations among all dimensions since multiple feasible channels and less important ones should be emphasized and filtered at the same time.
	Thus, a gating function is defined to calculate activated channel weight $ s_d $ as follows:
	\begin{equation}
	\label{equ:channel_weight}
	s_{d}
	=
	\mathrm{max} (z_{d} - t, 0) + \beta_t
	\ ,
	\end{equation}
	\noindent where $ \beta_t $ is the activation threshold for all weights. 
	Finally, the output of the dimensional attention strategy is obtained by resigning response of each channel $ R_d $ with $ s_{d} $ as follows: 
	\begin{equation}
	\label{equ:dim_act}
	R_{d}'
	=
	\sum\limits_{d=1}^{D} s_d R_{d}
	\ .
	\end{equation}
	
	As shown in \autoref{fig:main}, dimensions with higher reliability are given higher weights, while the ones with lower reliability cannot be activated. Finally, the peak and noise in the refined response can be enhanced and suppressed respectively, resulting in more accurate location.
	
	\subsection{Spatiotemporal attention strategy}
	\label{subsec:pa}
	
	Apart from the dimensional attention strategy, an element-wise multiplication with a predefined Hanning window and the pixel-wise sum of response map along the dimension axis is operated before normalization to the range $ [0, 1] $ and mean-subtraction.
	For each pixel $ s_{(i,j)} $ more than 0, the value will be activated with an exponential function, which indicates higher importance. 
	Thus, the static spatial attention map $ S $ is defined as follows:
	\begin{equation}
	\label{static_pos_map}
	S
	=
	\exp 
	\left[
	\mathrm{norm} 
	\left(
	\sum\limits_{d = 1}^{D} R_d \odot \mathbf{w}
	\right)
	\right]
	\ .
	\end{equation}
	
	Besides, the spatial information from the object motion is also taken into account. 
	Based on the current target size $ (X, Y) $ and the object position changes $ (\delta_{x}, \delta_{y}) $ caused by object motion or UAV viewpoint change in previous frame, the motion factor is calculated as follows:
	\begin{equation}
	\label{mot_factor}
	\gamma_t
	=
	\gamma
	\sqrt{
		\frac{\delta_{x}^{2} + \delta_{y}^{2}}
		{{X}^{2} + {Y}^{2}}
	}
	\ .
	\end{equation}
	
	With the shifting operation $\Delta_{x, y}$ derived from object motion, the dynamic attention map can be computed as:
	\begin{equation}
	\label{equ:dynamic_pos_map}
	S_d
	=
	S + \gamma_t S \Delta_{x, y}
	\ .
	\end{equation}

	Finally, a matrix multiplication to obtain the output is performed on the original response map as $R'' = S_d \odot R$.
	
	\begin{algorithm}[!t]
		\small
		\caption{TACF tracker}
		\label{alg:TACFtracker}
		\KwIn {Frames of the video sequence: $I_1, \cdots, I_K$.\\
			\hspace{0.35cm} The interval for context learning: $ f_c $.\\
			\hspace{0.35cm} The number of context patches: $S$.\\
			\hspace{0.35cm} Initialize the TACF in the first frame.}
		\KwOut {Predicted location in frame ${k}$.}
		\For{$k=2$ to end}{
			Extract the object patch in the frame $k$ from center location of the object in last frame\\
			Represent $\mathbf{x}_{n0}$ using  hand-crafted features\\        
			Enhance each channel of response maps with dimensional attention operation by \autoref{equ:dim_act}\\
			Evaluate the object motion and generate the spatiotemporal attention map before fusing each response maps by \autoref{equ:dynamic_pos_map}\\
			Calculate the location transformation in frame $k$ by searching the peak on the response maps\\
			\uIf{$\mod(k, f_c) == 0$}{
				Extract $ S $ context patches around the object\\
				\textbf{foreach} {context patch $\mathbf{x}_{ns}$} \textbf{do}\\
				\quad {Represent extracted patches using hand-crafted features and calculate penalty factor $p_s$ by \autoref{equ:penalty_factor}\\
					Learn new object appearance and update the model $ \hat{\mathbf{w}}_k^{\rm{(model)}} $ by \autoref{equ:model_result}}
			}
			\Else{
				Learn new object appearance and update the model $ \hat{\mathbf{w}}_k^{\rm{(model)}} $ by \autoref{equ:model_result_no_context}\\
			}
		}    
	\end{algorithm} 
	
	\section{Experiments}
	\label{sec:experi}
	
	In this section, the proposed TACF tracker is thoroughly evaluated on two well-known UAV tracking benchmarks with 173 challenging image sequences, \ie UAV123@10fps~\cite{Mueller2016ECCV} and UAVDT~\cite{Du2018ECCV}.
	
	\subsection{Experimental setups}
	
	\subsubsection{\textbf{Implementation details}}
	
	Our TACF is implemented with Matlab 2018a, and all the experiments are evaluated on a PC equipped with Intel i7-8700K CPU (3.7GHz) and NVIDIA GeForce RTX 2080 GPU.
	The TACF tracker employs two hand-crafted features, \ie histograms of gradients~\cite{Dalal2005CVPR} and color names~\cite{Danelljan2014CVPR}, to represent object and context patches. 
	The regularization parameter $ \lambda_1 $ and $ \lambda_2 $ is set to $ 5 \times 10^{-5} $ and $ 0.0625 $, respectively.
	The interval for context learning $ f_c $ and the number of context patches $S$ are set as 2 and 4.
	Details of the TACF tracker can be seen in \Autoref{alg:TACFtracker}.
	All parameters are fixed for all the experiments.
	
\begin{figure}[!t]
		\setlength{\abovecaptionskip}{-2pt}
		\centering
		\includegraphics[width=0.45\textwidth]{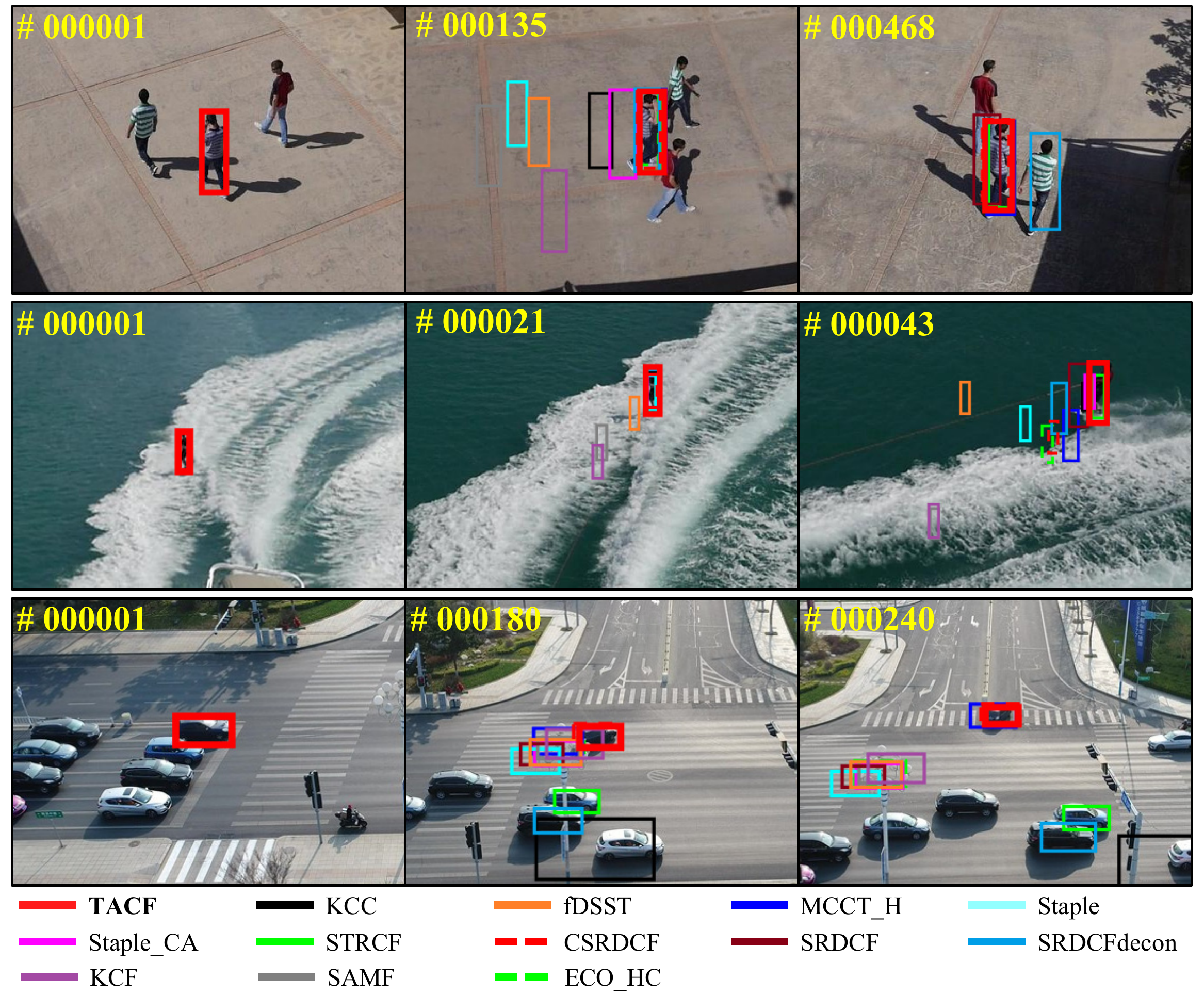}
		\caption{Examples of UAV tracking results. The first, second, and third row show the \textit{group1\_}3, \textit{wakeboard7} from UAV123@10fps benchmark, and \textit{S1606} video sequence from UAVDT benchmark.}
		\label{fig:result}
	\end{figure}

	Besides, \autoref{fig:result} shows some qualitative tracking results of TACF with the other 12 trackers in challenging UAV video sequences.
	Related source code and UAV tracking video are available in \url{https://github.com/vision4robotics/TACF-Tracker} and 
	\url{https://youtu.be/4IWKLmRoS38}.
	
	\subsubsection{\textbf{Benchmarks and evaluation methodology}}
	
	To validate the effectiveness of the proposed method, extensive experiments on UAV123@10fps~\cite{Mueller2016ECCV} and UAVDT~\cite{Du2018ECCV} benchmarks are conducted to evaluate overall performance.
	
	As the first comprehensive UAV tracking benchmark, UAV123@10fps contains 123 video sequences with more than 37K frames and 12 kinds of challenging visual attributes, making it the most significant object tracking benchmark from an aerial viewpoint.
	UAVDT benchmark focuses on complex scenarios with 50 representative video sequences, which are fully annotated with bounding boxes with up to 9 kinds of challenging visual attributes.
	
	Success rate (SR) is employed to evaluate the performance of the proposed tracker. 
	SR based on the one-pass evaluation protocol can illustrate the percentage of frames when the overlap ratio between the estimated and the ground truth bounding boxes is higher than a certain threshold. Following~\cite{Wu2015PAMI}, the area under curve (AUC) is adopted to rank the success rate of each tracker.
	
	\begin{figure*}[!ht]
		\setlength{\abovecaptionskip}{-2pt}
		\centering
		\subfigure{
			\begin{minipage}[t]{0.46\textwidth}
				\centering
				\includegraphics[width=1\textwidth]{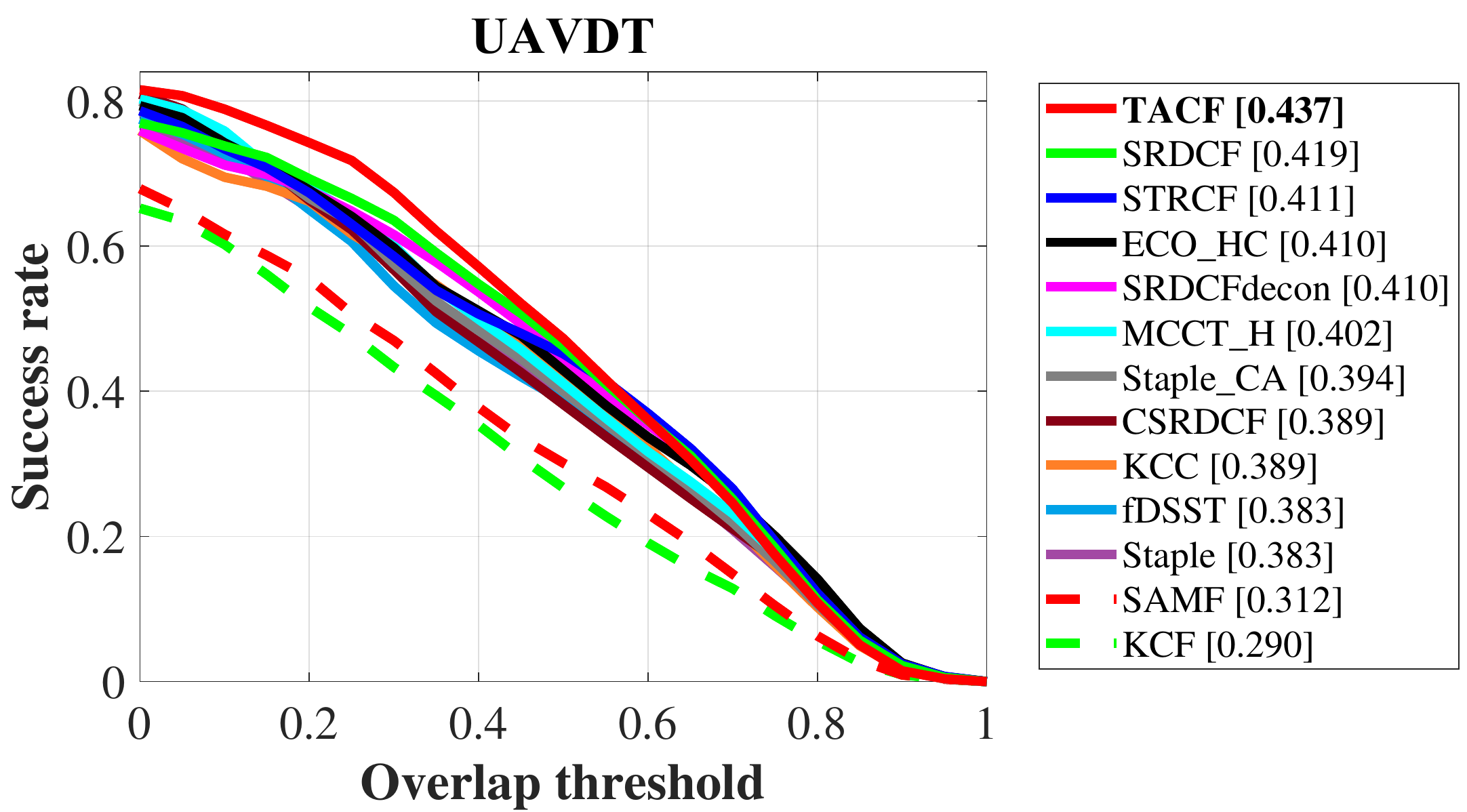}
			\end{minipage}
		}%
	\quad
		\subfigure{
			\begin{minipage}[t]{0.46\textwidth}
				\centering
				\includegraphics[width=1\textwidth]{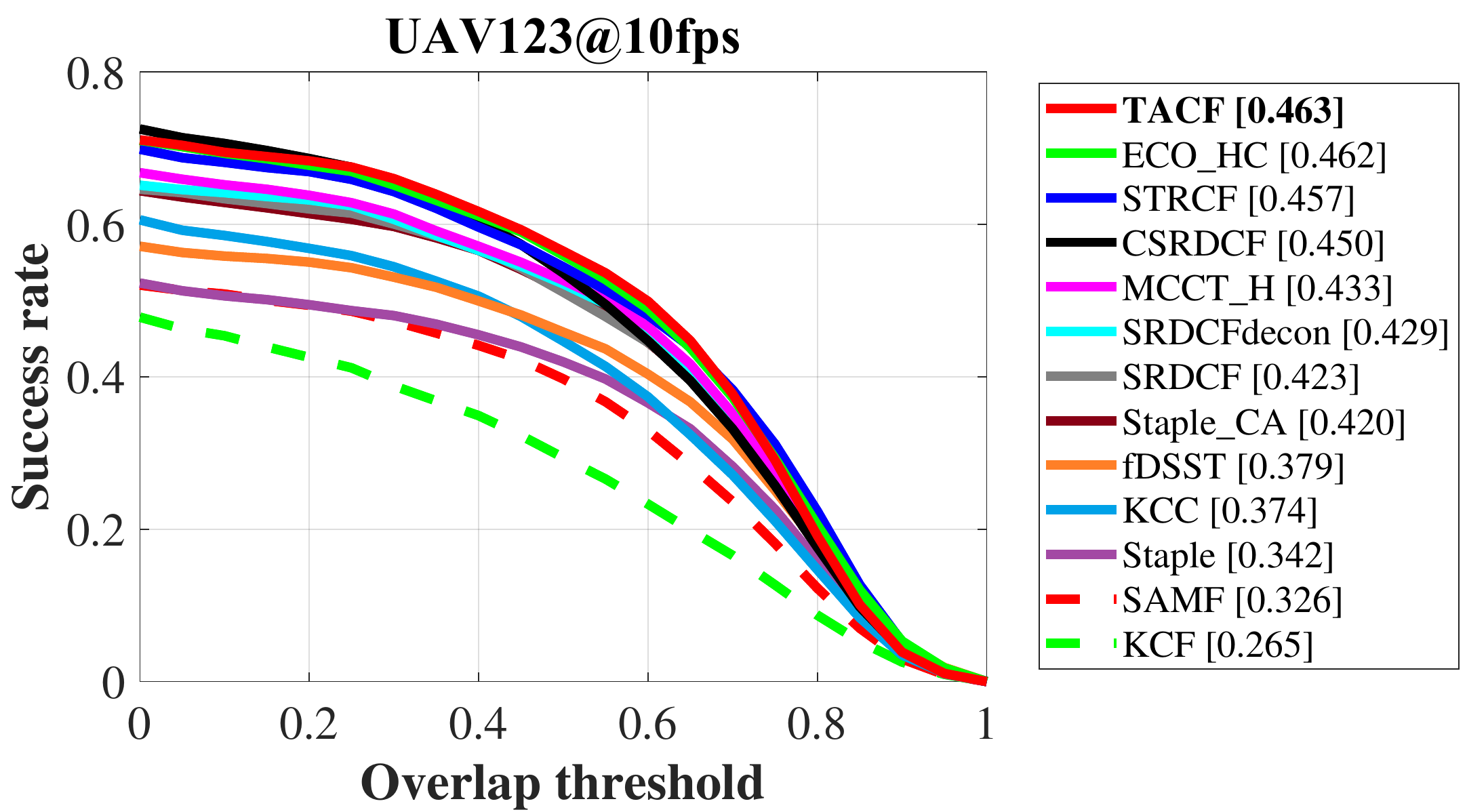}
			\end{minipage}
		}%
		\caption{Success plots of the proposed TACF tracker and other 12 state-of-the-art trackers on UAVDT and UAV123@10fps benchmarks. The experimental results demonstrate that our method yields superior performance on both widely-adopted UAV object tracking benchmarks.}
		\label{fig:SPs}
		\vspace{-4pt}
	\end{figure*}
	
	\begin{table*}[htbp]
		\scriptsize
		\centering
		\caption{The average FPS of TACF versus other state-of-the-art trackers on two well-known UAV object tracking benchmarks.}
		\label{tab:speed}%
		\begin{tabular}{ccccccccccccc|c}
			\toprule
			& KCF   & SAMF  & Staple & KCC   & fDSST & Staple\_CA & SRDCF & SRDCF\_decon & MCCT\_H & CSRDCF & STRCF & ECO\_HC & \textbf{TACF} \\
			\midrule
			Avg. FPS & 651.1 & 12.8  & 65.4  & 48.9  & 168.1 & 58.9  & 14    & 7.5   & 59.7  & 12.1  & 28.5  & 69.3  & \textbf{28.1} \\
			\bottomrule
		\end{tabular}%
	\vspace{-4pt}
	\end{table*}%
	
	\begin{figure*}[!ht]
		\setlength{\abovecaptionskip}{-4pt}
		\centering
		\subfigure{
			\begin{minipage}[t]{0.46\textwidth}
				\centering
				\includegraphics[width=1\textwidth]{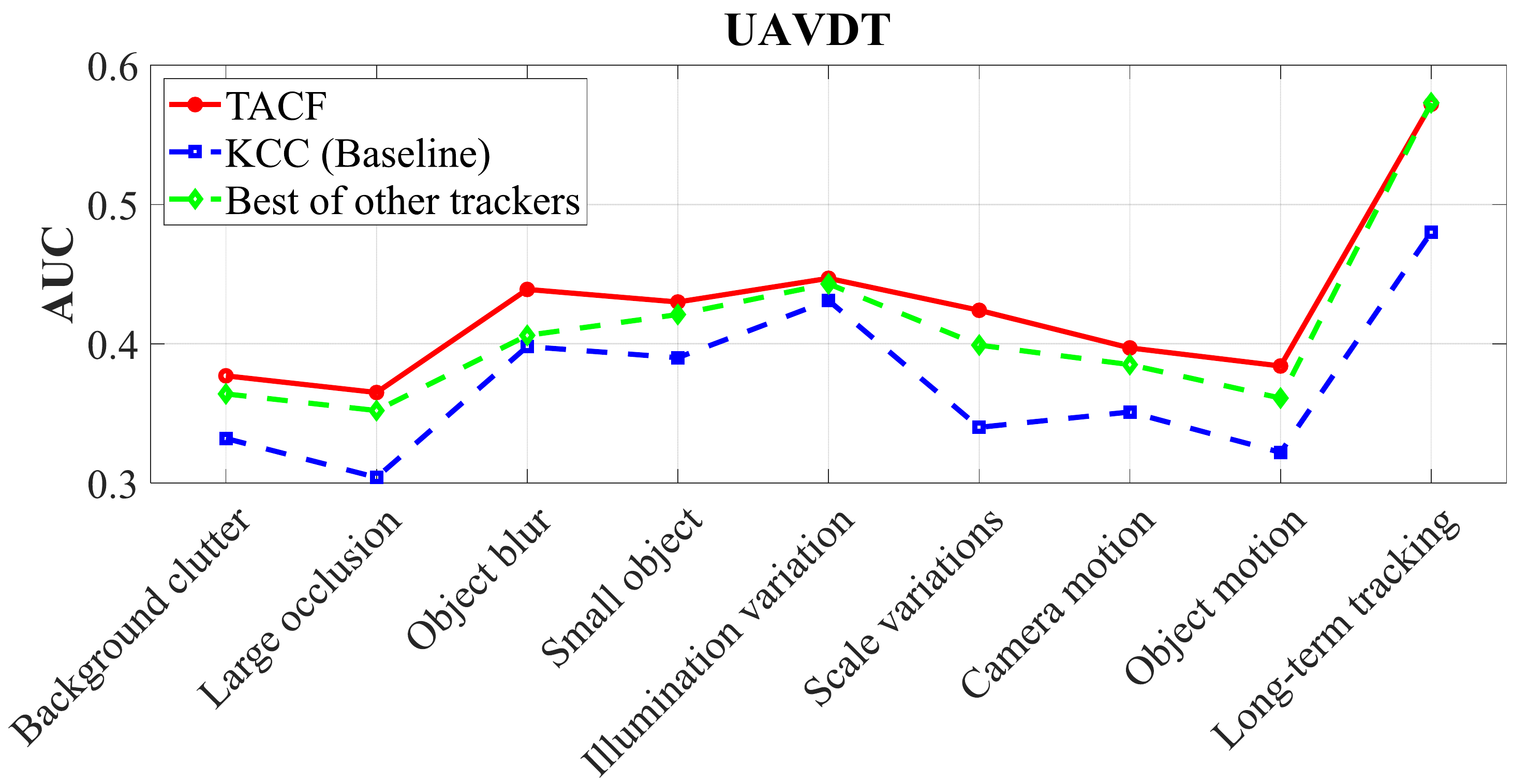}
			\end{minipage}
		}%
		\subfigure{
			\begin{minipage}[t]{0.46\textwidth}
				\centering
				\includegraphics[width=1\textwidth]{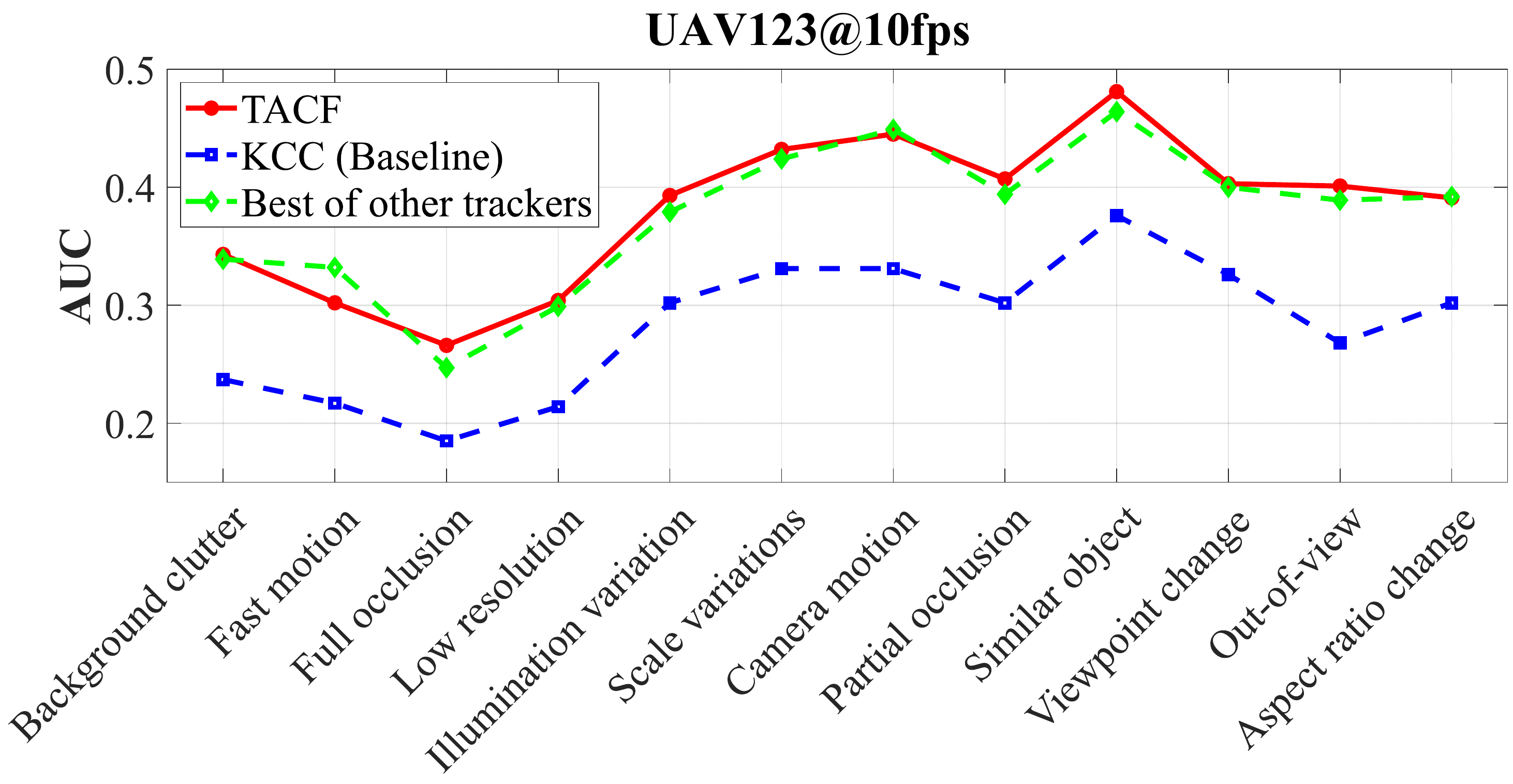}
			\end{minipage}
		}%
		\caption{Attribute-based evaluation between TACF and other state-of-the-art trackers on UAVDT and UAV123@10fps datasets. The AUC scores of different trackers in the success plots are applied to rank trackers.
		In most cases, the proposed TACF tracker performs favorably against other trackers, achieving significant improvements compared to the baseline tracker.}
		\label{fig:attribute_comp}
		\vspace{-4pt} 
	\end{figure*}
	
	\subsection{Qualitative experiments}
	\subsubsection{\textbf{State-of-the-art Comparison}}
	
	In real-world tasks, efficient operations of the tracking algorithm is particularly important due to the limited computing resources on UAV platform.
	Thus, the TACF tracker is compared with other 12 state-of-the-art trackers with hand-crafted features, including 
	MCCT\_H~\cite{Wang2018CVPR}, 
	Staple\_CA~\cite{Mueller2017CVPR}, 
	SRDCF~\cite{Danelljan2015ICCV}, 
	BACF~\cite{Galoogahi2017CVPR}, 
	KCC~\cite{Wang2018AAAI}, 
	CSRDCF~\cite{Lukezic2017CVPR}, 
	SAMF~\cite{Li2014ECCVws}, 
	Staple~\cite{Bertinetto2016CVPR}, 
	KCF~\cite{Henriques2015TPAMI}, 
	SRDCFdecon~\cite{Danelljan2016CVPR}, 
	STRCF~\cite{Li2018CVPR}, 
	fDSST~\cite{Danelljan2014BMVC},
	and ECO\_HC~\cite{Danelljan2017CVPR}.
	The open-source codes of trackers with default parameters provided by the authors are used in the following evaluations.
	
	As shown in \autoref{fig:SPs}, the proposed TACF tracker achieves better performance compared with other trackers in success plots.
	On the UAVDT dataset, TACF achieves the best score with 0.437, exceeding the second (SRDCF, 0.419) and third-best tracker (STRCF, 0.411) by $4.30\%$ and $6.32\%$, respectively.
	On the UAV123@10fps dataset, TACF keeps the best score, outperforming the second (ECO\_HC, 0.462) and third-best (STRCF, 0.457) trackers.
	
	In addition to excellent tracking performance, the speed of the proposed TACF tracker (28.1 FPS) is sufficient for UAV real-time tracking, as shown in \autoref{tab:speed}.
	Despite that KCF obtains the best tracking speed (651.1 FPS), followed by fDSST (168.1 FPS) and DSST (106.5 FPS), their tracking performance much lower than TACF.
	
	\subsubsection{\textbf{Attribute-based performance analysis}}
	
	The performance of the TACF tracker and other trackers are also analyzed in different attributes.
	\autoref{fig:attribute_comp} shows the scores of different trackers on different challenging attributes and demonstrates TACF exhibits better performance than most of the other trackers except for fast motion.
	Especially when partial occlusion or background clutter occurs, the proposed TACF tracker has a significant improvement over its baseline, and have achieved state-of-the-art performance in these aspects on these two benchmarks. 
	Usually, in a cluttered background, most CF-based methods tend to learn appearance models from both objects and irrelative noise.
	By applying the tri-attention strategy, CF can focus on crucial aspects so that the tracker can show better performance against these complex visual factors.
	
	In the future, it is possible to employ a mobile CNN to extract convolutional features for object representation to replace hand-crafted features with limited encoding ability.
	Features from different layers of CNN can provide semantic information and ensure efficient operation so that the tracker can improve the performance against object fast motion.
	
	\begin{figure}[!t]
		\setlength{\abovecaptionskip}{-2pt}
		\centering
		\includegraphics[width=0.45\textwidth]{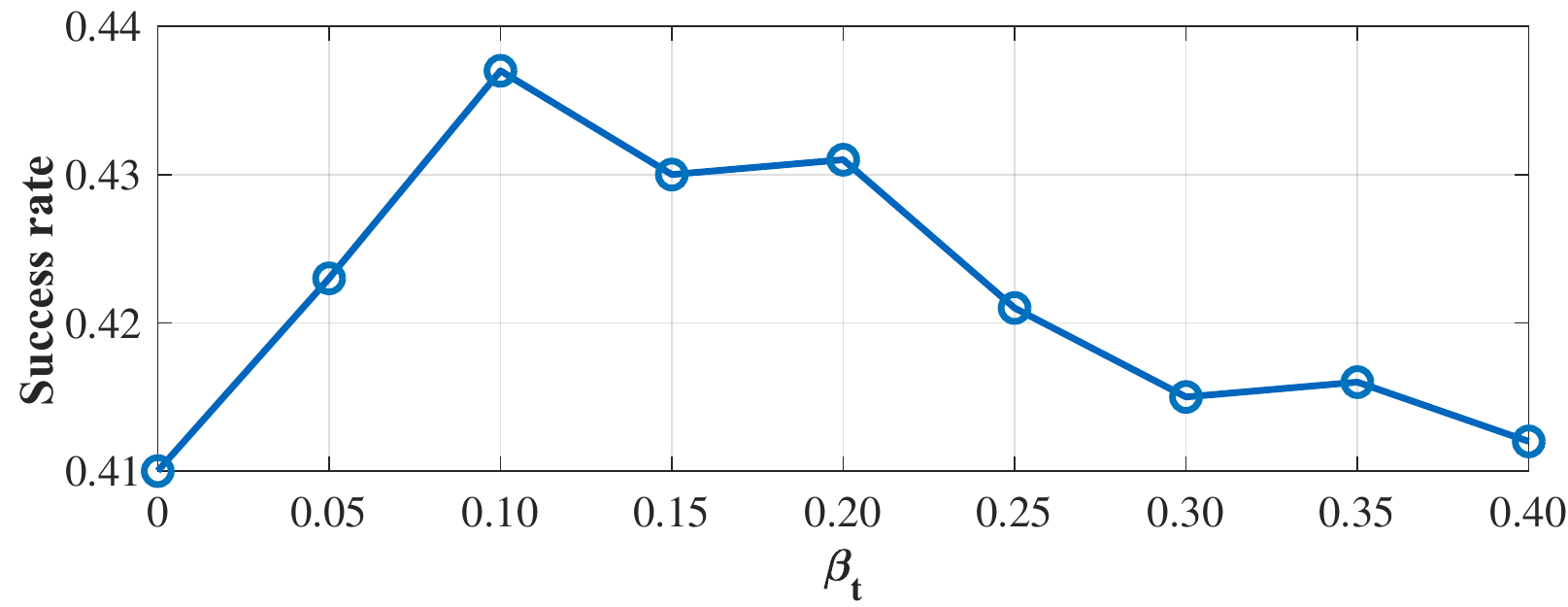} 
		\caption{Effect of activation threshold $ \beta_t $ on the UAVDT benchmark. When $ \beta_t > 0$ , it contributes to the overall performance of TACF tracker.}
		\label{fig:keyparam}
	\end{figure} 
	
	\subsubsection{\textbf{Key parameter analysis}}
	
	To verify the effectiveness of the activation threshold of dimensional attention modules on the tracking performance, different $ \beta_t $ is further analyzed on the UAVDT dataset. Starting from 0, $ \beta_t $ increases in small increments of 0.05 until 0.4.
	
	The activation threshold $ \beta_t $ determines the degree of focus heightened by the dimensional attention strategy: The higher the $ \beta_t $ is set, the more attention is paid to the relatively more important dimensions. TACF with $ \beta_t = 0 $ means that the filters
	trained with the consideration of all noisy dimensions. As shown in \autoref{fig:keyparam}, TACF effectively improves the success rate when $ \beta_t $ is set over 0. 
	The success rate reaches the peak (0.437) at $ \beta_t = 0.1 $. Thus, $ \beta_t = 0.1 $ is chosen for the best performance on the challenging UAV video sequences.
	
	\begin{table}[!b]
		\centering
		\caption{Performance comparisons between TACF with different modules on UAVDT and UAV@10fps benchmarks.}
		\label{tab:ablation}
		\begin{tabular}{ccccccc}
			\toprule
			\multirow{2}[4]{*}{Tracker} & \multicolumn{3}{c}{Module} & \multicolumn{2}{c}{Success rate} & \multirow{2}[4]{*}{MSPF} \\
			\cmidrule{2-6}          & CA    & DA    & SA    & UAVDT & UAV@10fps &  \\
			\midrule
			KCC     & $\xmark$ & $\xmark$ & $\xmark$ & 0.389 & 0.374 & 20.47 \\
			TACF+SA & $\xmark$ & $\xmark$ & $\checkmark$ & 0.423 & 0.398 & 20.98 \\
			TACF+DA & $\xmark$ & $\checkmark$ & $\xmark$ & 0.425 & 0.407 & 24.25 \\
			TACF+CA & $\checkmark$ & $\xmark$ & $\xmark$ & 0.432 & 0.421 & 32.25 \\
			\midrule
			\textbf{TACF}  & $\checkmark$ & $\checkmark$ & $\checkmark$ & 0.437 & 0.456 & 35.59 \\
			\bottomrule
		\end{tabular}%
	\end{table}%
	
	\subsubsection{\textbf{Ablation study}}
	
	Here in-depth analysis related to three different modules in the TACF framework, including contextual attention module (CA), dimensional attention (DA), and spatiotemporal attention module (PA), as well as the baseline tracker is performed to verify the effectiveness. Apart from the success rate, milliseconds per frame (MSPF) is applied to evaluate the average operational time cost.
	
	The baseline tracker, KCC, is considered as a particular case of TACF without the tri-attention strategy. 
	The $\checkmark$ in each column denotes that the corresponding module is activated while the $\xmark$ means that the module is deactivated.
	\autoref{tab:ablation} presents that the TACF tracker has significantly superior performance to KCC for $21.1\%$ and $8.4\%$ on UAV123@10fps and UAVDT dataset, respectively.
	Besides, three different attention strategy integrated into the original tracker has shown satisfying improvements from the baseline.
	
	In terms of the average time cost, each stage of TACF is operated within an acceptable time, \ie the average running time of CA, SA, and DA is about 11.48, 3.78, and 0.51 ms, respectively. As a result, the TACF tracker operates at an average speed of 28.1 FPS on a single CPU.
	
	Although the superior accuracy and robustness of TACF come at the cost of extra computational time, 
	it still can meet the real-time performance requirement of UAV object tracking applications. 
	On the one hand, due to the introduction of contextual attention, the processing speed for large object targets is reduced. 
	On the other hand, the current TACF tracker is implemented in MATLAB without additional engineering optimization. 
	Thus, appropriate parallel computing methods operated in the onboard processors can accelerate the operation speed when the proposed method is applied to real-world tracking tasks.
	
	\section{Conclusions}
	\label{sec:conclusion}
	In this work, a novel tracking framework with tri-attention correlation filters for robust UAV object tracking is proposed to achieve high performance in tracking tasks by leveraging multi attention mechanisms. Three types of attention, \ie contextual, spatiotemporal, and dimensional attention, have been effectively fused into the training and detection stages. Compared with the baseline tracker, the proposed TACF tracker has dramatically improved performance under challenging factors such as partial occlusion and background clutter. Moreover, qualitative and quantitative experiments on two established UAV tracking benchmarks demonstrate that the presented TACF tracker has outperformed other 12 state-of-the-art trackers in terms of accuracy, robustness, and efficiency. We believe, with our proposed tri-attention strategy, the correlation filters with multi-level attention can achieve better tracking performance further, and open the door to more extensive applications and researches for UAV.
	
	
	\section*{ACKNOWLEDGMENT}
	The work was supported by the National Natural Science Foundation of China (No.61806148).
	
	\bibliographystyle{IEEEtran}
	\bibliography{IROS2020}
	
\end{document}